\documentclass[]{spie}
\usepackage[]{graphicx}
\usepackage{bmpsize}
\usepackage{caption}
\usepackage{subcaption}
\usepackage{setspace}
\usepackage{placeins}
\usepackage{overpic}
\usepackage{wrapfig}
\usepackage{cite}
\usepackage{color}
\usepackage{multirow}
\usepackage{amsmath}
\definecolor{CadetBlue}{cmyk}{0.62,0.57,0.23,0}
\definecolor{redd}{rgb}{.6,.1,.1}
\definecolor{umass}{cmyk}{0,1,.63,.29}
\definecolor{MyDarkGreen}{rgb}{0.1,0.55,0} 
\usepackage[pdftex,colorlinks=true,linktocpage=true,pagebackref=false,linkcolor=blue,
citecolor=MyDarkGreen, bookmarks=true,pdfauthor={John B. Sigman},
bookmarksnumbered=true]{hyperref}
\DeclareCaptionFormat{spie}{\fontsize{9}{9}\selectfont#1#2#3}
\usepackage[printonlyused,withpage,nolist]{acronym}
\usepackage[marginpar]{todo}

\makeatletter
\AtBeginDocument{%
  \renewcommand*{\AC@hyperlink}[2]{%
    \begingroup
    \hypersetup{hidelinks}%
    \hyperlink{#1}{#2}%
    \endgroup
  }%
}
\makeatother

\makeatletter
\def\endthebibliography{%
  \def\@noitemerr{\@latex@warning{Empty `thebibliography' environment}}%
  \endlist
}
\makeatother

\title{Background Adaptive Faster R-CNN for Semi-Supervised Convolutional Object Detection of Threats in X-Ray Images}

\author{John B. Sigman, Gregory P. Spell, Kevin J Liang, and Lawrence Carin
  \skiplinehalf
  Duke University, Durham, NC 27708, United States of America\\
}

\authorinfo{Further author information:\\
  John B. Sigman, E-mail: john.sigman@gmail.com
}

\newcommand{\ProjectString}{HSTS04-16-C-CT7020}
\newcommand{\TSAProjectName}{\acs{TSA} Contract \ProjectString}
\newcommand{\naive}{na{\"i}ve}
\newcommand{\eg}{{\it{e.g.}}}
\newcommand{\ie}{{\it{i.e.}}}

\newcommand{\DatasetB}{Dataset B}
\newcommand{\DatasetA}{Dataset A}

\begin{document}
\graphicspath{{ARTWORK/}}
\maketitle 

% List of acronyms
\begin{acronym}[CCCCCCCC]
  \acro{AUC}{area under the curve}
  \acro{BA Faster R-CNN}{Background Adaptive Faster R-CNN}
  \acro{CNN}{Convolutional Neural Network}
  \acro{UDA}{Unsupervised Domain Adaptation}
  \acro{Faster R-CNN}{Faster Regions with Convolutional Neural Networks}
  \acro{IOU}{Intersection Over Union}
  \acro{LAGs}{liquids, aerosols, and gels}
  \acro{GRL}{Gradient Reversal Layer}
  \acro{mAP}{mean Average Precision}
  \acro{ROC}{Receiver Operating Characteristic}
  \acro{DANN}{Domain Adaptive Neural Network}
  \acro{SOC}{Stream-of-Commerce}
  \acro{ROI}{Region of Interest}
  \acro{TIP}{threat image projection}
  \acro{TSA}{Transportation Security Administration}
  \acro{AP}{Average Precision}
  \acro{$T_p$}{True Positive}
  \acro{$F_p$}{False Positive}
  \acro{$F_n$}{False Negative}
\end{acronym}

%%%%%%%%%%%%%%%%%%%%%%%%%%%%%%%%%%%%%%%%%%%%%%%%%%%% 

% -------ABSTRACT------%
\begin{abstract}
  Recently, progress has been made in the supervised training of Convolutional Object Detectors (\eg{} Faster R-CNN) for threat recognition in carry-on luggage using X-ray images. 
  This is part of the Transportation Security Administration's (TSA's) mission to ensure safety for air travelers in the United States.
  Collecting more data reliably improves performance for this class of deep algorithm, but requires time and money to produce training data with threats staged in realistic contexts. 
  In contrast to these hand-collected data containing threats, data from the real-world, known as the Stream-of-Commerce (SOC), can be collected quickly with minimal cost; while technically unlabeled, in this work we make a practical assumption that these are without threat objects. 
  Because of these data constraints, we will use both labeled and unlabeled sources of data for the automatic threat recognition problem. 
  In this paper, we present a semi-supervised approach for this problem which we call Background Adaptive Faster R-CNN. 
  This approach is a training method for two-stage object detectors which uses Domain Adaptation methods from the field of deep learning.
  The data sources described earlier are considered two ``domains'': one a hand-collected data domain of images with threats, and the other a real-world domain of images assumed without threats. 
  Two domain discriminators, one for discriminating object proposals and one for image features, are adversarially trained to prevent encoding domain-specific information. 
  Penalizing this encoding is important because otherwise the Convolutional Neural Network (CNN) can learn to distinguish images from the two sources based on superficial characteristics, and minimize a purely supervised loss function without improving its ability to recognize objects. 
  For the hand-collected data, only object proposals and image features completely outside of areas corresponding to ground truth object bounding boxes (background) are used. 
  The losses for these domain-adaptive discriminators are added to the Faster R-CNN losses of images from both domains. 
  This technique enables threat recognition based on examples from the labeled data, and can reduce false alarm rates by matching the statistics of extracted features on the hand-collected backgrounds to that of the real world data. 
  Performance improvements are demonstrated on two independently-collected datasets of labeled threats.
\end{abstract}
\acresetall

%%%%%%%%%%%%%%%%%%%%%%%%%%%%%%%%%%%%%%%%%%%%%%%%%%%% 
\section{Introduction}

Personal baggage security checkpoints consist of X-ray scanners and human operators, and their purpose is to prevent harm by capturing threatening objects and weapons.
Modern X-ray scanner systems use sophisticated technology to construct internal views of bags or belongings, but these systems are still reliant on human operators to identify and locate threats.
Bags can be highly cluttered environments, and owing to the transmissive nature of X-ray sensing, objects layer on top of each other in views.
Some examples of threat objects in bags can be seen in Figure \ref{fig:introduction-sample-detections}.
Operators must be vigilant to catch all of a diverse, constantly evolving, but relatively rare set of prohibited items; all while maintaining high passenger throughput. 
In order to reduce the cognitive load on these human threat screeners, we seek automated solutions for threat screening.

\newcommand{\introsamplefigurewidth}{.24\linewidth}
\begin{figure}[t]
  \centering
  \begin{subfigure}[b]{\introsamplefigurewidth}
    \includegraphics[width=\linewidth]{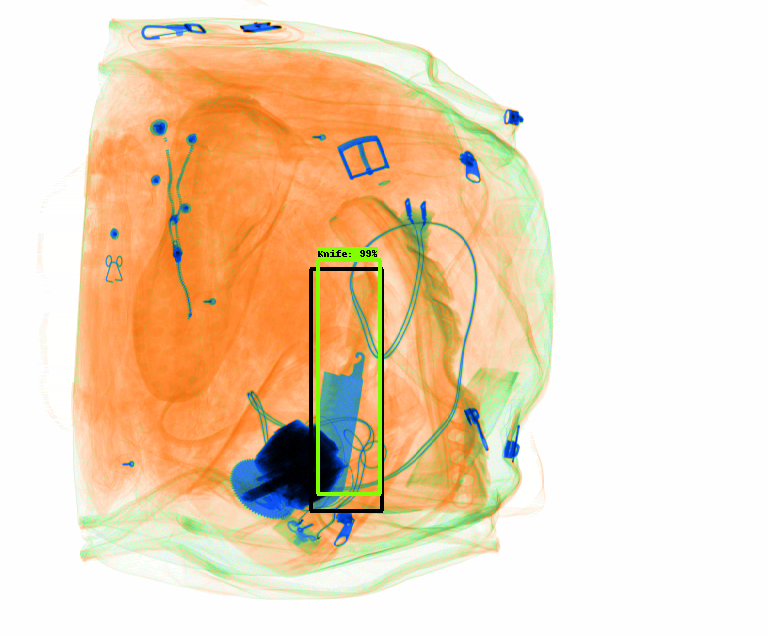}
    \caption{}
    \label{subfig:sample-detection-knife}
  \end{subfigure}
  \begin{subfigure}[b]{\introsamplefigurewidth}
    \includegraphics[width=\linewidth]{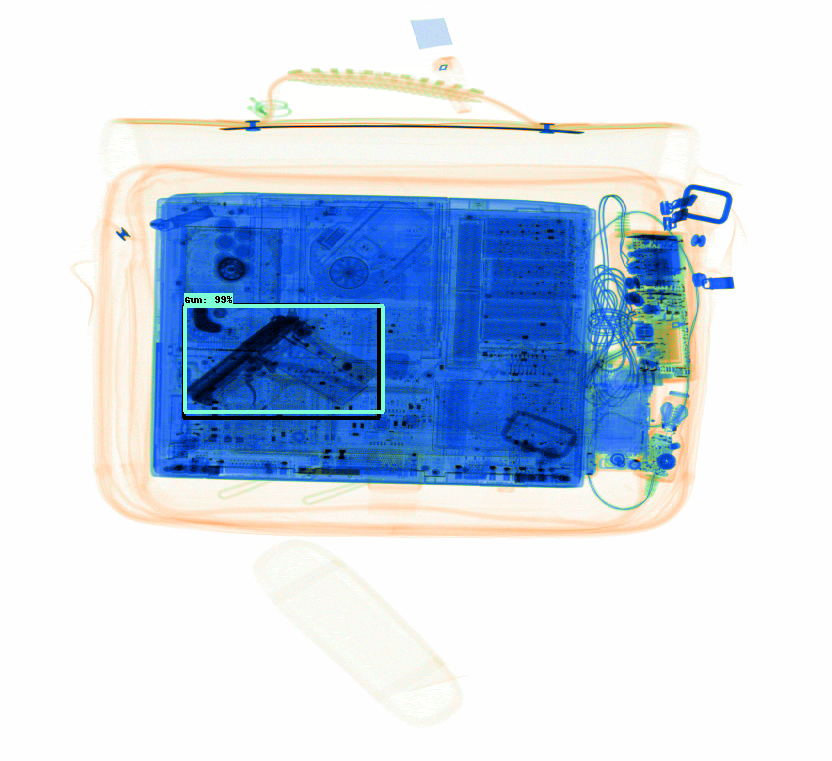}
    \caption{}
    \label{subfig:sample-detection-gun}
  \end{subfigure}
  \begin{subfigure}[b]{\introsamplefigurewidth}
    \includegraphics[width=\linewidth,trim={0 6cm 0 1cm}]{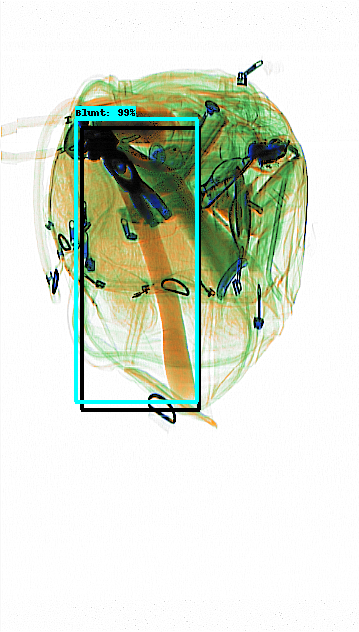}
    \caption{}
    \label{subfig:sample-detection-blunt}
  \end{subfigure}
  \begin{subfigure}[b]{\introsamplefigurewidth}
    \includegraphics[width=\linewidth,trim={0 4cm 0 1cm}]{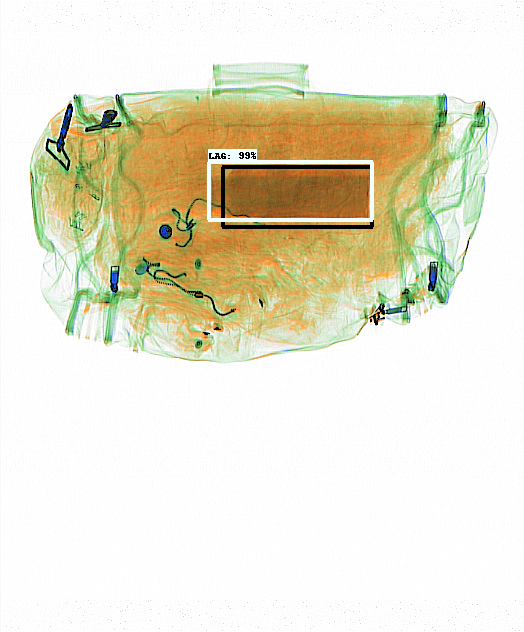}
    \caption{}
    \label{subfig:sample-detection-lag}
  \end{subfigure}
  \caption{Sample detections from mixed datasets of (\subref{subfig:sample-detection-knife}) knife, (\subref{subfig:sample-detection-gun}) gun, (\subref{subfig:sample-detection-blunt}) blunt, and (\subref{subfig:sample-detection-lag}) \acf{LAGs} threats. Images scanned with a laboratory prototype not in TSA configuration.}
  \label{fig:introduction-sample-detections}
\end{figure}

In the field of computer vision, identifying and localizing objects in a scene is called object detection.
Recently, convolutional neural networks (CNNs)~\cite{Lecun1989} have resulted in major improvements in performance, with many modern object detection models~\cite{Girshick2014,Girshick2015,liu2016ssd,Ren2015,Redmon2016} utilizing CNNs as a key component.
Given their excellent performance on benchmark datasets such as Cityscapes~\cite{Cordts2016} and MS COCO~\cite{Lin2014}, these models are now being considered for certain real world settings, including baggage screening at airport checkpoints.
However, the baggage scan images are different from the kinds of images in common object detection benchmarks.
In particular, while common benchmark datasets are constructed to have a high number of object instances per image, threat items at checkpoints occur comparably infrequently.
For example, the United States Transportation Security Administration (TSA) caught 4,432 firearms in carry-on baggage at US airport checkpoints in 2019, but against a backdrop of a billion passengers~\cite{TSA2020}.

This paper concerns an application of convolutional object detection using unlabeled imagery to improve threat detection in X-ray images.
Unlike the expensive staging, imaging, and labeling of threat data, real-world scans can be readily and cheaply collected from functioning airport checkpoints.
These real-world data are referred to as the \ac{SOC}.
Modern deep neural network techniques are extremely data hungry, showing log-linear improvements with dataset size, \ie{} roughly linear performance increases with exponentially increasing numbers of examples~\cite{Sun2017}.
A technique which could train with these vast quantities of unlabeled images to improve threat detection is therefore highly valuable.
Using such images would not be intended to improve recall (otherwise known as probability of detection).
Instead, by exposing the model during training to a broader variety of real-world backgrounds we could reduce false alarm rates.
However, strictly supervised use of unlabeled imagery as negative examples (the \ac{SOC}) contains risks.
Distinguishing patterns in the data (images) between the \ac{SOC} and the labeled dataset could be discovered by a sufficiently flexible neural network, resulting in brittle predictions due to the difference in object occurrences between these two sets.
In our case, this means that, if used \naive{ly}, the network can recognize domain differences between hand-labeled data and \ac{SOC} data.
Once the model has learned to distinguish the domains, it can encode that images from the \ac{SOC} all contain no objects and suppress detections, undermining the goal of real-world threat detection.
In order to overcome this capacity of a \ac{CNN} to distinguish domains, we will use techniques from domain adaptation~\cite{Ganin2015, Li2019}.

In this paper, our contribution is to demonstrate the real-world use of a variant of Domain Adaptive Faster R-CNN~\cite{Chen2018} that accounts for target shift between the image domains.
Our Background Adaptive Faster R-CNN makes a strong assumption about the prior probability of targets in the target domain, particularly that images from this domain have no foreground (threat) objects.
First we will give a brief introduction to \ac{UDA} and its use in object detection.
Then, we will describe some related works in the field of automatic threat detection in X-ray images, focusing on recent work in modern deep convolutional approaches.
Next, we will describe how our method is used to enable learning from the threatless real-world data (the \ac{SOC}).
In the results section, we will show some improvements in performance using our technique to a baseline \ac{Faster R-CNN}~\cite{Ren2015} approach.
These positive results demonstrate successful use of the SOC data and hold great promise for future applications of automatic threat recognition to airport security checkpoints.

%%%%%%%%%%%%%%%%%%%%%%%%%%%%%%%%%%%%%%%%%%%%%%%%%%%% 

\newcommand{\RealWorldTextSignifier}{\textrm{SOC}}
\newcommand{\HandCollectedTextSignifier}{\textrm{HC}}

\newcommand{\RealWorldDistribution}{p^{\RealWorldTextSignifier}}
\newcommand{\HandCollectedDistribution}{p^{\HandCollectedTextSignifier}}

\newcommand{\RealWorldJointDistribution}{\RealWorldDistribution (x,y)}
\newcommand{\RealWorldEmptyObjectPriorProbability}{\RealWorldDistribution (|y| > 0)}
\newcommand{\RealWorldObjectPriorProbability}{\RealWorldDistribution (y)}%

\newcommand{\HandCollectedJointDistribution}{\HandCollectedDistribution (x,y)}
\newcommand{\HandCollectedEmptyObjectPriorProbability}{\HandCollectedDistribution (|y| > 0)}
\newcommand{\HandCollectedObjectPriorProbability}{\HandCollectedDistribution (y)}

\newcommand{\RealWorldDataset}{\mathcal D^{\RealWorldTextSignifier}}
\newcommand{\HandCollectedDataset}{\mathcal D^{\HandCollectedTextSignifier}}

\section{Background}
\label{sec:background}
In this section, we will describe the concepts and theory from recent works of machine learning which are applied in \ac{BA Faster R-CNN}.
Specifically, we will introduce the concepts of adversarial training,  target shift, and covariate shift from the \ac{UDA} literature.

\subsection{Unsupervised Domain Adaptation}

\newcommand{\TargetDataset}{$\mathcal D^T$}
\newcommand{\SourceDataset}{$\mathcal D^S$}
% Domain adaptation is a recent popular approach for learning from two datasets from different distributions.
In the most basic formulation of adversarial domain adaptation~\cite{Ganin2015, Goodfellow2014}, we have access to a labeled dataset \SourceDataset{} from a source domain and an unlabeled dataset \TargetDataset{} from a target domain, and we wish to train a classifier that performs well on test data from the target domain without using any labels from \TargetDataset.
We will consider two types of shifts between data domains, defining them now as they pertain to image classification~\cite{Li2019,redko2018}.
First, target shift refers to unequal label prior probabilities between two distributions, $p(y) \neq p'(y)$.
The other type is covariate shift, which refers to a difference between the conditional  distributions of an image given its class category: $p(x|y) \neq p'(x|y)$.
One way to improve performance on the target dataset is to train a domain-invariant classifier through matching the marginal distribution of features extracted from the data $x$ of either dataset \TargetDataset{} or \SourceDataset{}.
The marginal distribution of features, and not a class-specific conditional feature distribution, is used because it only requires the data and not the labels.
\ac{DANN} \cite{Ganin2015} introduced a method for learning domain-invariant features by using an adversarial loss via domain discriminator.
If the marginal distribution of features $h$ extracted from target data are referred to by $p^{T}(h)$, and the marginal distribution of features extracted from source data are $p^{S}(h)$, then the domain-invariant features are achieved when:
\begin{align}
  p^{S}(h) = p^{T}(h)
  \label{eq:da-matching-marginal-features}
\end{align}
Because these features depend on learnable parameters in a \ac{CNN}, stating that probability distributions are equal is not a static description of the data, but implies the matching of these distribution by learning a suitable feature extractor using adversarial training~\cite{Chen2018, Ganin2015, Goodfellow2014}.
As shown previously~\cite{Li2019}, if the prior probabilities of the target domain classes were known, $p^{T}(y)$, then an adapted feature extractor for the target domain can be learned by matching the distributions:
\begin{align}
  p^{T}(h) = q^{S}(h) = \sum_{c}p^{S}(h|y=c)p^{T}(y=c)
  \label{eq:da-matching-marginal-features}
\end{align}
Where $c$ is one of the classes in \TargetDataset{} or \SourceDataset{}, and $q^{S}$ is a model distribution which is the class-conditional weighted average of source features, weighted according to their incidence in the target domain.
This is not done under normal conditions because \textit{unsupervised} domain adaptation implies that we do not have $p^{T}(y=c)$.

\subsection{Unsupervised Domain Adaptation in Object Detection}
These concepts were extended to object detection in a self-driving car setting~\cite{Chen2018} by matching the distributions of CNN features of the overall images and the proposed objects, demonstrating that an object detection model trained on mostly clear weather day images can still perform well on night, foggy, or inclement weather settings.
As in \ac{DANN}, the method assumed that only marginal feature distributions needed to be matched to produce domain-invariant features.
In this case, this was a safe assumption, because the target dataset of cloudy images was synthetically produced by transforming the source dataset of sunny images, guaranteeing equal likelihood of object occurrences in images from both sets.
This simple treatment of the object occurrences in the two data sources will not work for threat recognition in X-ray images.
An individual image from the \ac{SOC} has a low prior probability of containing a threat while labeled data were all staged with threats.
%%%%%%%%%%%%%%%%%%%%%%%%%%%%%%%%%%%%%%%%%%%%%%%%%%%% 

\section{Related Works}
This paper describes an application of recent developments in computer vision to the problem of automated threat detection in luggage, but this problem has been studied for many years.
Early work using machine learning for X-ray image classification relied on hand-crafted features such as Difference of Gaussians (DoG) and scale-invariant feature transform (SIFT) \cite{lowe1999sift}. 
These hand-crafted features were then fed to a traditional classifier such as a Support Vector Machine (SVM)~\cite{cortes1995support}, leveraging an approach known as Bag-of-Visual-Words \cite{Bastan2011, Bastan2013, Turcsany2013, Mery2016, Kundegorski2016}.

The first application of deep learning algorithms to images of X-ray baggage involved manually cropping regions of x-ray images and classifying the crops according to different categories of firearms and knives, as well as classes of camera and laptop \cite{akcay2016transfer}. This was accomplished via transfer learning, in which a pre-trained \ac{CNN} was fine-tuned for the specific datasets of X-ray baggage.  Deep object detection algorithms have also been investigated for use in X-ray baggage scans \cite{akcay2017evaluation, akcay2018using, Liang2018, Liang2019}. In these investigations, a range of \ac{CNN} feature extractors have been examined, including VGG \cite{Simonyan2015}, Inception V2 \cite{inception_v2}, and ResNet \cite{he2016deep}. Furthermore, a variety of \ac{CNN}-based detection algorithms are adapted for X-ray baggage: \ac{Faster R-CNN} \cite{Ren2015}, R-FCN \cite{Dai2016}, and YOLOv2 \cite{Redmon2017}. We encourage readers to see a recent survey \cite{Akcay2020} for a more thorough overview of these efforts.  A subset of these works \cite{Liang2018, Liang2019} specifically sought to incorporate deep learning techniques into the security systems used by the \ac{TSA} at U.S. airport checkpoints. 
In many settings, acquiring unlabeled data is significantly easier than labeled data.
For X-Ray baggage scanner threat detection settings, this is especially the case, as acquiring labeled data often also requires assembling the threat-containing bags and scanning them.
Thus, semi-supervised approaches have drawn considerable interest as an efficient way to leverage more data.
Another approach is \ac{TIP}, which digitally adds threat objects into a bag using various synthetic methods~\cite{Bhowmik2019}, this is possible here due to the transmissive nature of X-ray, .

\section{Methods} \label{sec:methods}
\subsection{Data}

Consider two data-generating processes $\RealWorldJointDistribution$ and $\HandCollectedJointDistribution$, which model the appearance of objects, $y=\{(c,b)\}$, in an image $x$.  
In this notation, the set $y$ contains objects described by $c$, a categorical random variable indicating class, and $b \in \mathbb{R}^4$  describing the coordinates of the object's bounding box.
Let $\RealWorldDataset$ be the set of data samples from $\RealWorldJointDistribution$, and let $\HandCollectedDataset$ be the set of data samples from $\HandCollectedJointDistribution$.
Data from $\RealWorldJointDistribution$ are our \acf{SOC} data, gathered from checkpoints, and are the target domain.
Data from $\HandCollectedJointDistribution$ are our ``Hand-Collected'' source domain data, which were collected by subcontractors staging threats in realistic context during \TSAProjectName{}~\cite{Liang2018,Liang2019}. 

By design, $\RealWorldDistribution$ and $\HandCollectedDistribution$ have a target shift between them.
Namely, $\RealWorldDistribution$ contains little to no threat objects, $p^{SOC}(|y| > 0) = 0$, and every image in $\HandCollectedDistribution$ contains a threat object, $p^{HC}(|y| > 0) = 1$.
We can make the following statements of target shift (Equation \ref{eq:target-shift-in-data}) and covariate shift (Equation \ref{eq:covariate-shift-in-data}):
\begin{align}
  \RealWorldObjectPriorProbability & \neq \HandCollectedObjectPriorProbability \label{eq:target-shift-in-data}\\
  \RealWorldDistribution (x|y) & \neq \HandCollectedDistribution (x|y) \label{eq:covariate-shift-in-data}
\end{align}
We seek a method for learning an object detection model which addresses these disparities to more effectively find threats in the target domain.

\subsection{Supervised Auxiliary Negative Training}

\label{subsec:supervised-auxiliary-negative-training}
A \naive{} way to learn from the objectless images is to treat them identically as the labeled examples in the training set.
This implicitly treats all regions of the \ac{SOC} images as background~\cite{yang2020object}, and tends to not work in practice.
First, if the number of labeled images is small, adding too many unlabeled images dilutes the dataset, resulting in a large class imbalance that may drown out the learning signal for positive samples. 
Down-weighting negative samples emphasizes the learning signal for the actual objects of interest, but also reduces the value of the unlabeled set.
Hard negative mining, or the focal loss \cite{Lin2017} focus learning on the hardest background examples, but makes the assumption that the unlabeled data is from the same distribution as labeled dataset.

In this specific case, $\RealWorldDataset$ and $\HandCollectedDataset$ have some inherent differences which we cast as covariate shift.
We hypothesize that these differences could include but not be limited to: idiosyncratic reconstruction noise on the periphery of images, signatures from the individual laboratory prototypes used to stage threat data, or effects from the close queuing of bags in scans which occurs in the real world and not in staged data collections.
While these might not seem like meaningful differences, a sufficiently expressive feature extractor can pick up on these discrepancies and rapidly learn to suppress any detections in \ac{SOC} images.
While this reduces false positives, it could reduce the model's effectiveness when detecting threats in the real world, because it was only ever shown real-world data without threats.

We have conducted this experiment and have observed empirically that the \ac{Faster R-CNN} model is able to shatter metrics on the train \textit{and test} \ac{SOC} images.
That is, while having some level of expected generalization error when evaluated on the test set of hand-labeled data, the model can predict without any false alarms across the test set of \ac{SOC} data.
These domain-dependent differences between the hand-collected data and the \ac{SOC} are not a problem when the object detection model is trained with only labeled data, possibly only slightly increasing generalization error.
However, they could be catastrophic if the model is allowed to learn to discriminate domains, which occurs when using supervised learning on negative examples from the \ac{SOC}.

\subsection{Adversarial Training and Background Matching}
\newcommand{\venndiagramfigurewidth}{.3\linewidth}
\begin{figure}[t]
  \centering
  \begin{subfigure}[b]{\venndiagramfigurewidth}
    \includegraphics[angle=-90,width=\linewidth]{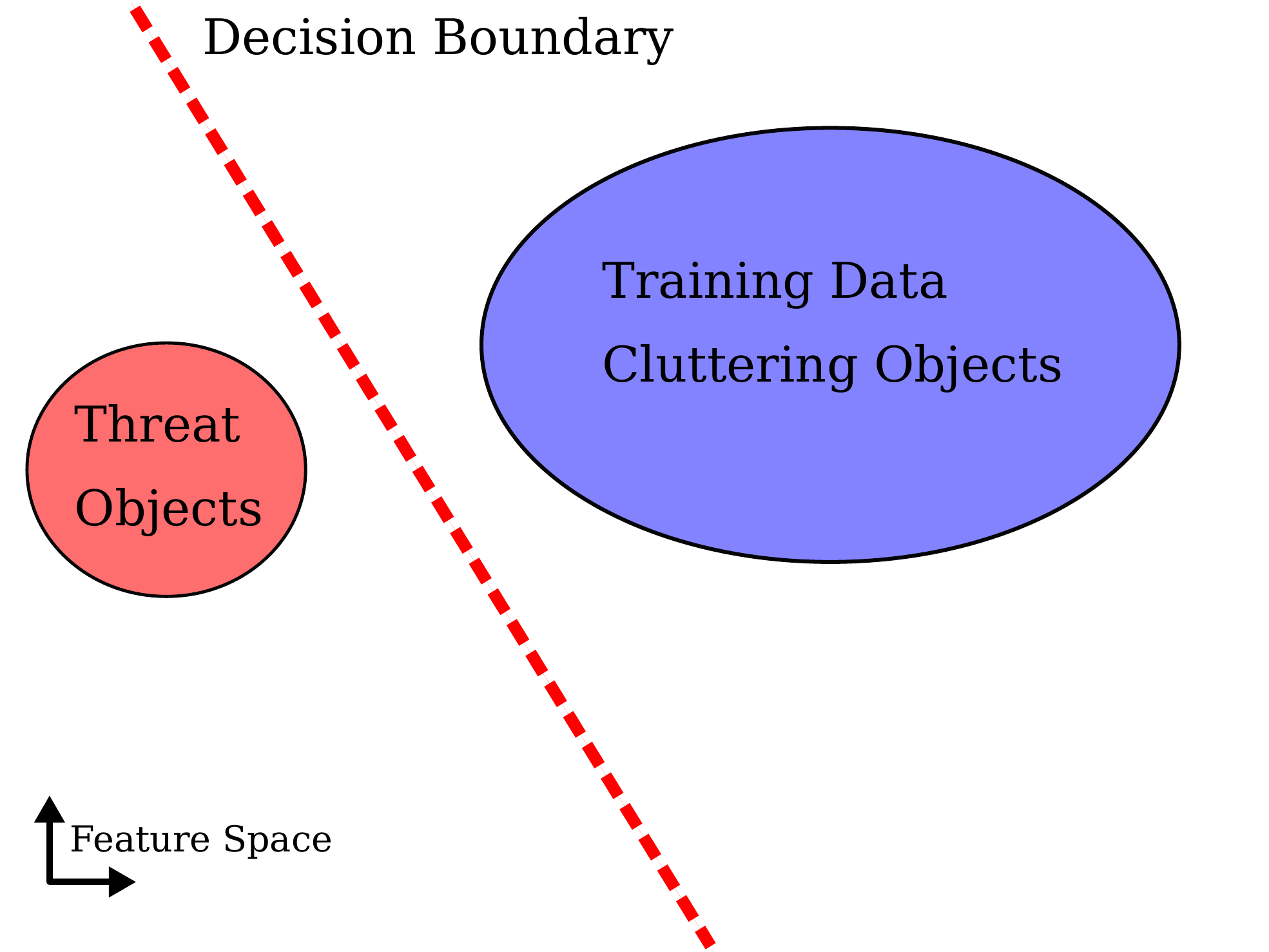}
    \caption{}
    \label{subfig:venn-diagram-supervised-feature-space}
  \end{subfigure}
  \begin{subfigure}[b]{\venndiagramfigurewidth}
    \includegraphics[angle=-90,width=\linewidth]{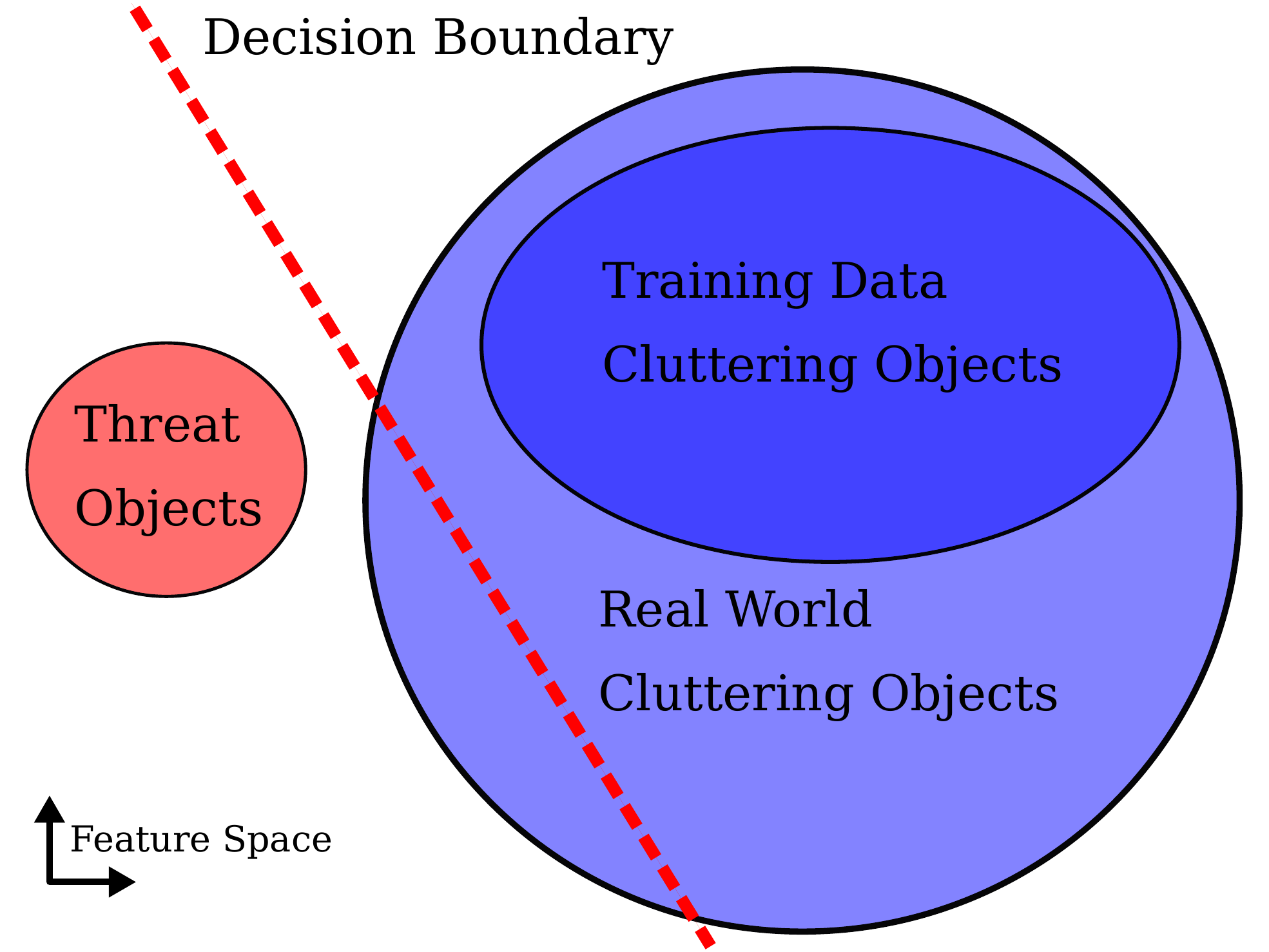}
    \caption{}
    \label{subfig:venn-diagram-supervised-plus-real-world}
  \end{subfigure}
  \begin{subfigure}[b]{\venndiagramfigurewidth}
    \includegraphics[angle=-90,width=\linewidth]{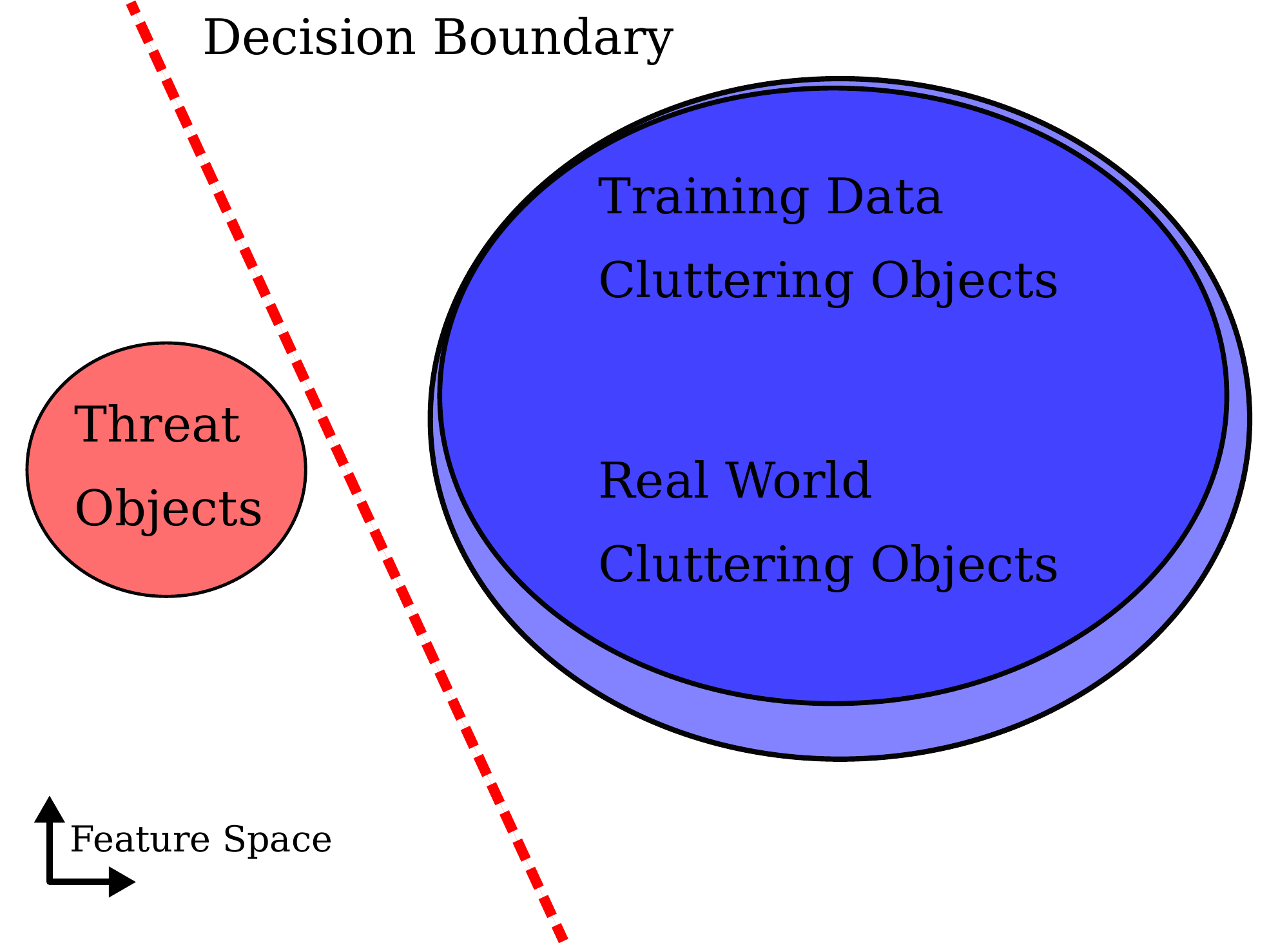}
    \caption{}
    \label{subfig:venn-diagram-semisupervised-feature-space}
  \end{subfigure}
  \caption{Conceptual diagram included to demonstrate the intuition of this approach for anomaly/threat detection. (\subref{subfig:venn-diagram-supervised-feature-space})~Separation of threats and background when training with labeled data. (\subref{subfig:venn-diagram-supervised-plus-real-world}) Application of the supervised learned threat detector when applied to heterogenous real-world data. The set of ``Real World Cluttering Objects'' crosses over the decision boundary to imply  increased false alarm rates in the real world. (\subref{subfig:venn-diagram-semisupervised-feature-space}) Using adversarial training for the feature extractor to match the statistics of the backgrounds with the real-world data decreases false alarms while still learning robust features.}
  \label{fig:venn-diagram-figure}
\end{figure}

\label{subsub:methods-background-matching}
Our proposed method, \acf{BA Faster R-CNN}, is a training procedure applied to the \ac{Faster R-CNN} model~\cite{Ren2015}, which selectively applies the unsupervised domain adaptive Faster R-CNN~\cite{Chen2018} losses.
Its intended use is in an anomaly/threat detection situation, with two separate datasets, such as $\HandCollectedDataset$ and $\RealWorldDataset$.
Images from both sets train the Faster R-CNN loss for threat detection, and images from both are used simultaneously in adversarial domain adaptation.
The method exposes the model to a wider variety of backgrounds than those in $\HandCollectedDataset$ and through adversarial distribution matching, the CNN has barriers preventing it from encoding which dataset each object proposal comes from based on image signatures unrelated to the objects in a bag.
A conceptual diagram is shown in Figure \ref{fig:venn-diagram-figure}.

We will borrow previous notation~\cite{Chen2018} to describe probability distributions of features which are part of the \ac{Faster R-CNN} algorithm.
$p(B,I)$ is a marginal probability distribution of features extracted from feature map pixels $I$ within bounding box $B$.
$p(I)$ is a marginal probability distribution of the extracted feature map pixel, when each pixel is treated as an independent observation.

Extending previous work~\cite{Chen2018}, we address a particular case of target shift between domains in addition to covariate shift.
As described in Equation \ref{eq:da-matching-marginal-features}, if we know the prior probability of labels in the target domain, we can perform adversarial domain adaptation by matching target features with class-conditional source features, weighted according to their incidence in the target domain.
This requires matching the model distributions $q(B,I)$ and $q(I)$ to their respective distribution of features in the target $p^{SOC}(B,I)$ and $p^{SOC}(I)$:
\begin{align}
  p^{SOC}(B,I) = q(B,I) &= \sum_{c}{p^{HC}(B,I | c) p^{SOC}(c)} \label{eq:object-detection-full-matching-instances}\\
  p^{SOC}(I) = q(I) &= \sum_{c}{p^{HC}(I | c) p^{SOC}(c)}\label{eq:object-detection-full-matching-images}
\end{align}
The class-conditional distributions of instance features $p(B,I|c)$ and image features $p(I|c)$ refer to features corresponding to the locations of ground truth classes $c$, which can be one of many threat classes or background.
Under the assumption of no threats in the \ac{SOC}, the class prior probability is one for background and zero for any foreground class.
\begin{align}
  p^{SOC}(c)=\begin{cases}
          1 \quad & c = \textrm{background}\\
          0 \quad & \textrm{otherwise}
        \end{cases}
  \label{eq:piecewise-definition-of-class-prior-probability}
\end{align}
Substituting Equation \ref{eq:piecewise-definition-of-class-prior-probability} into Equations \ref{eq:object-detection-full-matching-instances} and \ref{eq:object-detection-full-matching-images} gives:
\begin{align}
  p^{SOC}(B,I) &= p^{HC}(B,I|c=\textrm{background})   \label{eq:instance-matching}\\
  p^{SOC}(I) &= p^{HC}(I|c=\textrm{background})  \label{eq:image-matching}
\end{align}
Despite a target shift between the two domains, we can still build from previous techniques~\cite{Chen2018} so long as we only match the background portions of the hand-collected data to the \ac{SOC}.
We achieve background instance matching (Equation \ref{eq:instance-matching}) by first obtaining candidate object proposals in the standard way via Faster R-CNN.  
For object proposals from $\HandCollectedDataset$, we designate as background any proposal for which the \ac{IOU} with a ground truth threat object is below a certain threshold (in our experiments, 0.01). 
These background proposals are fed to an adversarial domain discriminator for object instances, along with all proposals made from $\RealWorldDataset$.

For background image matching (Equation \ref{eq:image-matching}), we first anti-crop the features of images from $\HandCollectedDataset$ according to the ground truth boxes of threats.
These feature maps, which have been masked in pixels corresponding to ground truth box locations, are used to train the adversarial pixel domain discriminator, alongside all of the feature map pixels from $\RealWorldDataset$. A third domain adaptation loss, a consistency regularization~\cite{Chen2018}, enforces that the two domain classifiers predict the same domain. This loss is implemented as the $\ell_2$ distance between the domain classifier outputs for an image. 
%%%%%%%%%%%%%%%%%%%%%%%%%%%%%%%%%%%%%%%%%%%%%%%%%%%% 

\section{Results} \label{sec:results}
In this section we give results for the application of \ac{BA Faster R-CNN} to images of threats in baggage.
These results are not meant to be compared across datasets, only within them.
While each scan of a single bag produces multiple views (each an image), we calculate precision/recall scores and \ac{mAP} on a single-image basis, deliberately excluding the sizable impact that multiple views of a threat per bag has on detection rates for simplicity.
This impact is outside the scope of our paper; our purpose is to indicate results of applying \ac{BA Faster R-CNN} to the problem of threat detection in carry-on luggage.
While we show here some improvements in precision and recall of using \ac{BA Faster R-CNN} on the labeled data collected in \TSAProjectName{}, its real purpose is to improve generalization performance when tested on held-out images from the \ac{SOC}.
Due to national security concerns and data sensitivity, we omit these results.

\subsection{Datasets}
\label{subsec:datasets}
Results in Section \ref{sec:results} were trained with datasets of over 6,000 labeled images for \DatasetA{} and over 19,000 labeled images for \DatasetB{}.
A training set of over 35,000 SOC images were used for \DatasetA{} and over 70,000 \ac{SOC} images were used for \DatasetB{}.
Performance was evaluated with over 1,100 labeled images from each set, and 500 \ac{SOC} images from each set were used as a representative sample to estimate false alarm rates.

\subsection{Labeled Data Performance}
\label{subsec:labeled-results}
For our performance metrics, we will show improvements in the precision/recall metric on a held out labeled set of images from two different vendors, \DatasetA{} and \DatasetB{}.
For our experiments, we used the Tensorflow Object Detection implementation~\cite{Huang2017}, which we extended for X-ray images and object detection domain-adaptive components.
All experiments were conducted with ResNet101~\cite{he2016deep} as the feature extractor and \ac{Faster R-CNN}~\cite{Ren2015} as the meta-architecture.
For all experiments, we use domain adaptive loss weighting~\cite{Chen2018} $\lambda = 0.1$ and \ac{GRL} weight~\cite{Chen2018,Ganin2015} 0.1.

We ran three different experiments on each dataset.
First, we trained a supervised Faster R-CNN baseline.
Then, we included only the instance loss component, which matches feature distributions of extracted proposals as in Equation \ref{eq:instance-matching}.
Finally, we trained a model using instance losses and image losses.
Image losses are used to match the extracted feature pixel distributions, as in Equation \ref{eq:instance-matching}.
When both image and instance losses are used, we also include the consistency loss~\cite{Chen2018}.

\subsubsection{\DatasetA{}}
\begin{figure}[tbh]
  \centering
  \includegraphics[width=.5\textwidth]{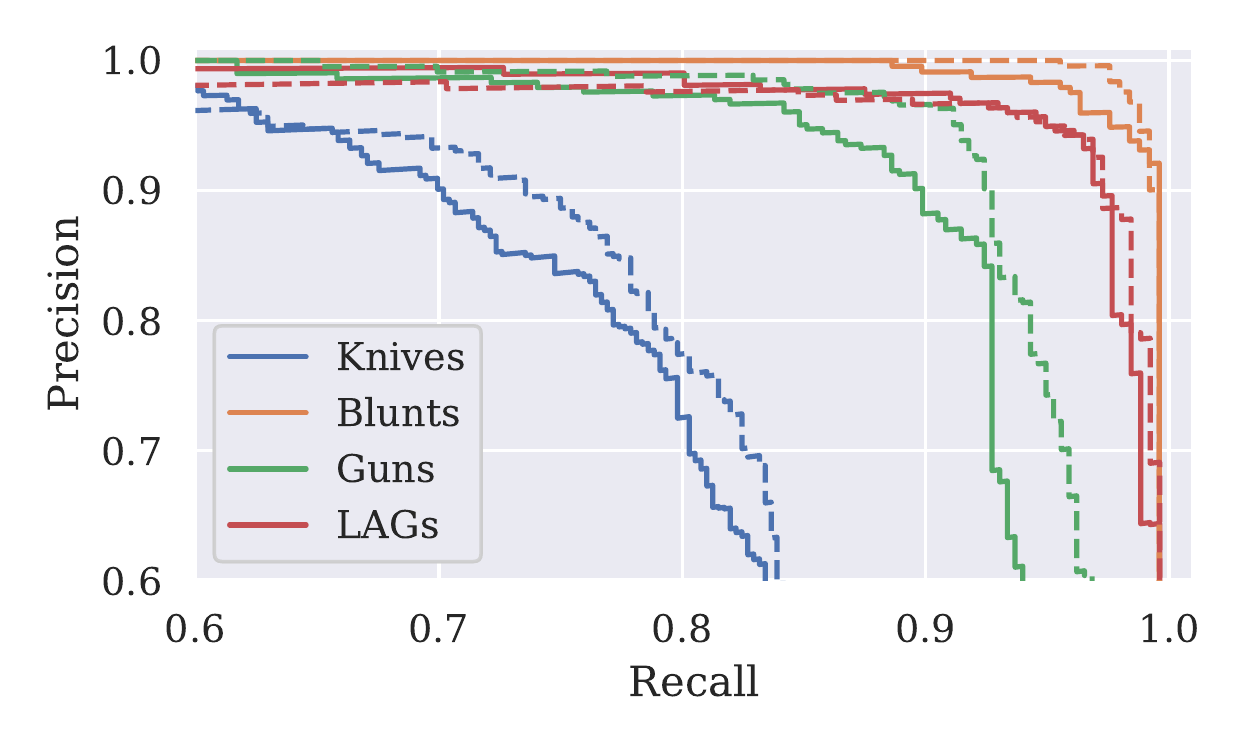}
  \caption{Precision/Recall for \DatasetA{}, treating all views as independent images. The solid lines correspond to the purely Supervised model, and the dashed lines correspond to Semi-Supervised use of \ac{BA Faster R-CNN} when all three domain-adaptive loss terms are applied. These results are summarized in Table \ref{table:a-labeled-precision-recall}}
  \label{fig:a-labeled-precision-recall}
\end{figure}

\DatasetA{} saw performance increases for nearly all labeled data classes when evaluated on the held-out set.
These precision/recall metrics are graphed in Figure \ref{fig:a-labeled-precision-recall}, and summarized in Table \ref{table:a-labeled-precision-recall}.
\ac{AP} on Knives data was only improved by using all three loss terms, while \ac{AP} was the best for the  \ac{LAGs} class when used with supervised learning.

\label{subsubsec:a-labeled-results}
% \begin{wrapfigure}{r}{0.5\textwidth}
% \end{wrapfigure}

\begin{table}[h]
  \centering
  \resizebox{\columnwidth}{!}{%
  \begin{tabular}{| l | c c c | c c c c |} 
    \hline
    & \multicolumn{3}{c |}{Loss term} & \multicolumn{4}{c|}{Threat Class}\\ \hline
    \multicolumn{1}{|c|}{Model Version}      & Instance & Image & Consistency & Knives & Blunts & Guns  & \acs{LAGs} \\ \hline
    \ac{Faster R-CNN}, Figure \ref{fig:a-labeled-precision-recall}:  (\textbf{\textemdash{}})   &               &            &                  & 0.832  & 0.989  & 0.931 & \textbf{0.980} \\ \hline
    +Match instances & \checkmark    &            &                  & 0.829  & 0.989  & 0.948 & 0.978 \\ \hline
    +Match instances and images, Figure \ref{fig:a-labeled-precision-recall}: (\textendash{} \textendash{})     & \checkmark    & \checkmark &    \checkmark    & \textbf{0.839}  & \textbf{0.992}  & \textbf{0.955} & 0.978 \\ \hline
  \end{tabular}
  }
  \caption{Tabulated \acp{AP} on labeled data from \DatasetA{}.}
  \label{table:a-labeled-precision-recall}
\end{table}

\FloatBarrier
\subsubsection{\DatasetB{}}
\begin{figure}[tbh]
  \centering
  \includegraphics[width=.5\textwidth]{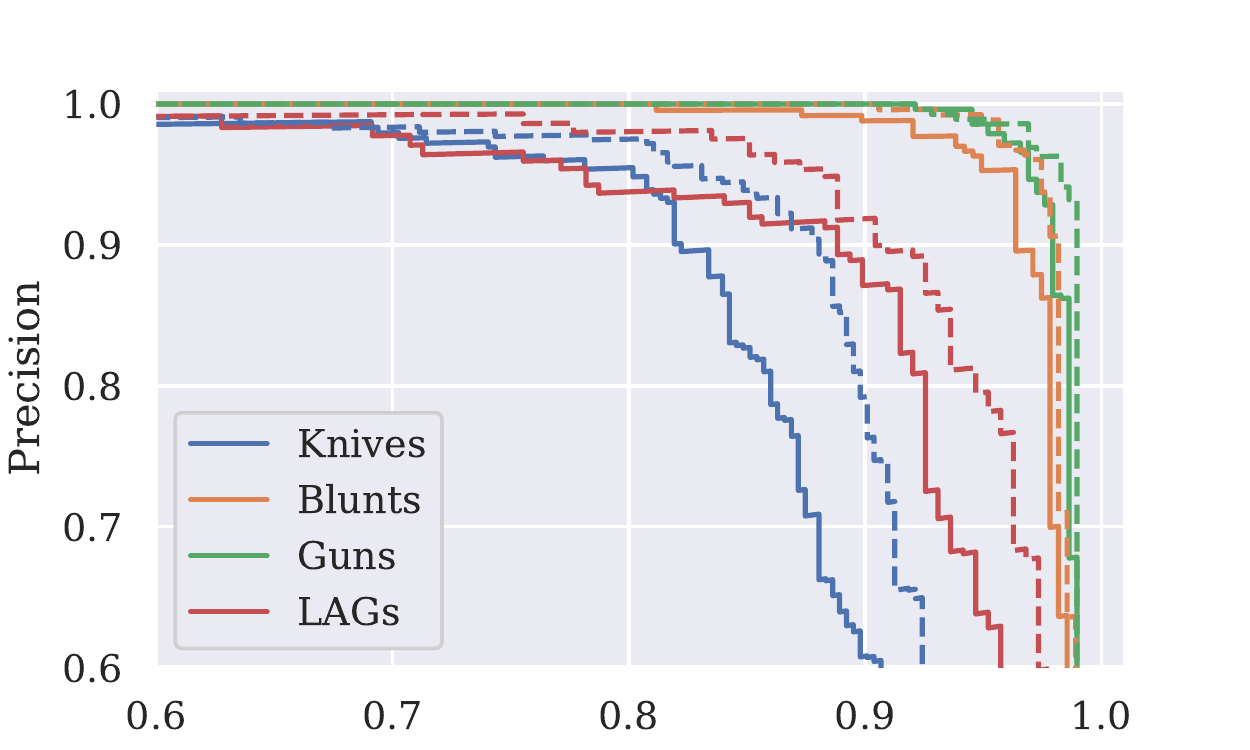}
  \caption{Precision/Recall for \DatasetB{}, treating all views as independent images. The solid lines correspond to the purely Supervised model, and the dashed lines correspond to Semi-Supervised use of \ac{BA Faster R-CNN} when all three domain-adaptive loss terms are applied. These results are summarized in Table \ref{table:b-labeled-precision-recall}}
  \label{fig:b-labeled-precision-recall}
\end{figure}

Figure \ref{fig:b-labeled-precision-recall} shows precision/recall curves for the labeled held-out set from \DatasetB{}.
There was an increase in the \ac{mAP} metric for all labeled data classes using background domain adaptation.
Guns and \ac{LAGs} did slightly worse using all three domain-adaptive losses than when only applied to instance losses.

\label{subsubsec:b-labeled-results}
% \begin{wrapfigure}{r}{0.5\textwidth}
% \end{wrapfigure}

\begin{table}[h]
  \centering
  \resizebox{\columnwidth}{!}{%
  \begin{tabular}{| l | c c c | c c c c |} 
    \hline
    & \multicolumn{3}{c |}{Loss term} & \multicolumn{4}{c|}{Threat Class}\\ \hline
    \multicolumn{1}{|c|}{Model Version} & Instance & Image & Consistency & Knives & Blunts & Guns  & \acs{LAGs} \\ \hline
    \ac{Faster R-CNN}, Figure \ref{fig:b-labeled-precision-recall}: (\textemdash{})         &               &            &                  & 0.894  & 0.977  & 0.985 & 0.936 \\ \hline
    +Match instances & \checkmark    &            &                  & 0.908  & 0.980  & \textbf{0.987} & \textbf{0.956} \\ \hline
    +Match instances and images, Figure \ref{fig:b-labeled-precision-recall}: (\textendash{} \textendash{}) & \checkmark    & \checkmark &    \checkmark    & \textbf{0.914}  & \textbf{0.982}  & 0.986 & 0.951 \\ \hline
  \end{tabular}	
  }
  \caption{Tabulated \acp{AP} on labeled data from \DatasetB{}.}
  \label{table:b-labeled-precision-recall}
\end{table}

\FloatBarrier

%%%%%%%%%%%%%%%%%%%%%%%%%%%%%%%%%%%%%%%%%%%%%%%%%%%% 

\section{Concluding Remarks}
This paper describes Background Adaptive Faster R-CNN, a technique for using real-world unlabeled data to improve threat detection in X-ray images.
As in a typical \acf{UDA} setup, two datasets are used.
The first of these is a source dataset, which is small and contains labeled examples of objects.
The second dataset is our target domain, and contains many more images without labels.
By matching the class conditional feature distributions of the background in the labeled data to the marginal distribution in the unlabeled data, we can use \ac{UDA} techniques despite target shift between the two datasets.

This technique was  required in order to use large quantities of real-world data because of the well-known tendency of deep convolutional feature extractors to learn brittle features if these are enough to separate the training data.
In our case, this meant that with supervised use of large quantities of real-world \ac{SOC} data, then the feature extractor need only recognize that the image was from the \ac{SOC} to achieve low losses.
Instead, we penalize the encoding of domain-specific information via two adversarial domain discriminators, which require the model learn from cluttering objects in the \ac{SOC}.

We demonstrate \ac{BA Faster R-CNN} on two independently-collected datasets. In this application, the target threat classes are guns, knives, blunt objects, and liquid, aerosols, or gels.
For the X-ray baggage datasets that we examine, we find that \ac{BA Faster R-CNN} succeeds in improving precision/recall performance for most threat classes in the labeled portion of the dataset. It is possible for \ac{BA Faster R-CNN} to reduce false alarm rates in the real world, but we omit such results here.

\subsection*{Acknowledgments}
This work was funded by the \acf{TSA} under Contract \#HSTS04-16-C-CT7020
We thank our sponsors in the Department of Homeland Security as well as our collaborators in the X-ray sensing industry.

% \subsection{Todos}
%\todos
% \clearpage
\bibliographystyle{spiebib}   %>>>> makes bibtex use spiebib.bst
\bibliography{refs}

\begin{thebibliography}{10}

\bibitem{Lecun1989}
Y.~Lecun, B.~Boser, J.~Denker, D.~Henderson, R.~E. Howard, W.~Hubbard, and
  L.~D. Jackel, ``{Backpropagation Applied to Handwritten Zip Code
  Recognition},'' {\em Neural Computation} , 1989.

\bibitem{Girshick2014}
R.~Girshick, J.~Donahue, T.~Darrell, and J.~Malik, ``{Rich feature hierarchies
  for accurate object detection and semantic segmentation},'' {\em Conference
  on Computer Vision and Pattern Recognition} , 2014.

\bibitem{Girshick2015}
R.~Girshick, ``{Fast R-CNN},'' {\em International Conference on Computer
  Vision} , 2015.

\bibitem{liu2016ssd}
W.~Liu, D.~Anguelov, D.~Erhan, C.~Szegedy, S.~Reed, C.-Y. Fu, and A.~C. Berg,
  ``{SSD: Single Shot MultiBox Detector},'' {\em European Conference on
  Computer Vision} , 2016.

\bibitem{Ren2015}
S.~Ren, K.~He, R.~Girshick, and J.~Sun, ``{Faster R-CNN: Towards Real-Time
  Object Detection with Region Proposal Networks},'' {\em Advances in Neural
  Information Processing Systems} , 2015.

\bibitem{Redmon2016}
J.~Redmon, S.~Divvala, R.~Girshick, and A.~Farhadi, ``{You Only Look Once:
  Unified, Real-Time Object Detection},'' {\em Conference on Computer Vision
  and Pattern Recognition} , 2016.

\bibitem{Cordts2016}
M.~Cordts, M.~Omran, S.~Ramos, T.~Rehfeld, M.~Enzweiler, R.~Benenson,
  U.~Franke, S.~Roth, and B.~Schiele, ``{The Cityscapes Dataset for Semantic
  Urban Scene Understanding},'' {\em Conference on Computer Vision and Pattern
  Recognition} , 2016.

\bibitem{Lin2014}
T.-Y. Lin, M.~Maire, S.~Belongie, L.~Bourdev, R.~Girshick, J.~Hays, P.~Perona,
  D.~Ramanan, C.~L. Zitnick, and P.~Dol{\'{i}}, ``{Microsoft COCO: Common
  Objects in Context},'' {\em European Conference on Computer Vision} , 2014.

\bibitem{TSA2020}
``{TSA Year in Review: 2019},'' {\em
  https://www.tsa.gov/blog/2020/01/15/tsa-year-review-2019} , 2020.

\bibitem{Sun2017}
C.~Sun, A.~Shrivastava, S.~Singh, and A.~Gupta, ``{Revisiting Unreasonable
  Effectiveness of Data in Deep Learning Era},'' {\em International Conference
  on Computer Vision} , 2017.

\bibitem{Ganin2015}
Y.~Ganin, E.~Ustinova, H.~Ajakan, P.~Germain, H.~Larochelle, F.~Laviolette,
  M.~Marchand, and V.~Lempitsky, ``{Domain-Adversarial Training of Neural
  Networks},'' {\em Journal of Machine Learning Research} , 2015.

\bibitem{Li2019}
Y.~Li, S.~Dai, L.~Carin, and D.~Carlson, ``{On Target Shift in Adversarial
  Domain Adaptation},'' {\em AISTATS} , 2019.

\bibitem{Chen2018}
Y.~Chen, W.~Li, C.~Sakaridis, D.~Dai, and L.~V. Gool, ``{Domain Adaptive Faster
  R-CNN for Object Detection in the Wild},'' {\em Conference on Computer Vision
  and Pattern Recognition} , 2018.

\bibitem{Goodfellow2014}
I.~J. Goodfellow, J.~Pouget-Abadie, M.~Mirza, B.~Xu, D.~Warde-Farley, S.~Ozair,
  A.~Courville, and Y.~Bengio, ``{Generative Adversarial Networks},'' {\em
  Advances In Neural Information Processing Systems} , 2014.

\bibitem{redko2018}
I.~Redko, N.~Courty, R.~Flamary, and D.~Tuia, ``Optimal transport for
  multi-source domain adaptation under target shift,'' {\em arXiv preprint
  arXiv:1803.04899} , 2018.

\bibitem{lowe1999sift}
D.~G. Lowe, ``{Object Recognition from Local Scale-Invariant Features},'' in
  {\em In Proceedings of the International Conference on Computer Vision
  (ICCV)},   {\bf 99}(2), pp.~1150--1157, 1999.

\bibitem{cortes1995support}
C.~Cortes and V.~Vapnik, ``Support-vector networks,'' {\em Machine
  learning}~{\bf 20}(3), pp.~273--297, 1995.

\bibitem{Bastan2011}
M.~Ba{\c{s}}tan, M.~R. Yousefi, and T.~M. Breuel, ``{Visual Words on Baggage
  X-Ray Images},'' {\em Computer Analysis of Images and Patterns.} ,
  pp.~360--368, 2011.

\bibitem{Bastan2013}
M.~Bastan, W.~Byeon, and T.~Breuel, ``{Object Recognition in Multi-View Dual
  Energy X-ray Images},'' in {\em In Proceedings of the British Machine Vision
  Conference (BMVC)},  January 2013.

\bibitem{Turcsany2013}
D.~Turcsany, A.~Mouton, and T.~P. Breckon, ``{Improving Feature-Based Object
  Recognition for X-Ray Baggage Security Screening Using Primed Visual
  Words},'' in {\em In Proceedings of the IEEE International Conference on
  Industrial Technology (ICIT)},  February 2013.

\bibitem{Mery2016}
D.~Mery, E.~Svec, and M.~Arias, ``{Object Recognition in Baggage Inspection
  Using Adaptive Sparse Representations of X-ray Images},'' {\em Image and
  Video Technology} , pp.~709--720, 2016.

\bibitem{Kundegorski2016}
M.~E. Kundegorski, S.~Ak{\c{c}}ay, M.~Devereux, A.~Mouton, and T.~P. Breckon,
  ``{On Using Feature Descriptors as Visual Words for Object Detection within
  X-ray Baggage Security Screening},'' in {\em In Proceedings of the
  International Conference on Imaging for Crime Detection and Prevention
  (ICDP)},  November 2016.

\bibitem{akcay2016transfer}
S.~Ak{\c{c}}ay, M.~E. Kundegorski, M.~Devereux, and T.~P. Breckon, ``{Transfer
  Learning Using Convolutional Neural Networks for Object Classification within
  X-ray Baggage Security Imagery},'' in {\em In Proceedings of the IEEE
  International Conference on Image Processing (ICIP)},  September 2016.

\bibitem{akcay2017evaluation}
S.~Akcay and T.~P. Breckon, ``{An Evaluation of Region Based Object Detection
  Strategies within X-ray Baggage Security Imagery},'' in {\em In Proceedings
  of the IEEE International Conference on Image Processing (ICIP)},  September
  2017.

\bibitem{akcay2018using}
S.~Akcay, M.~E. Kundegorski, C.~G. Willcocks, and T.~P. Breckon, ``{Using Deep
  Convolutional Neural Network Architectures for Object Classification and
  Detection within X-ray Baggage Security Imagery},'' {\em IEEE Trans. Info.
  Forens. Sec.}~{\bf 13}(9), pp.~2203--2215, 2018.
\newblock doi: 10.1109/TIFS.2018.2812196.

\bibitem{Liang2018}
K.~J. Liang, G.~Heilmann, C.~Gregory, S.~Diallo, D.~Carlson, G.~Spell,
  J.~Sigman, K.~Roe, and L.~Carin, ``{Automatic Threat Recognition of
  Prohibited Items at Aviation Checkpoints with X-Ray Imaging: a Deep Learning
  Approach},'' {\em Proc SPIE, Anomaly Detection and Imaging with X-Rays (ADIX)
  III} , 2018.

\bibitem{Liang2019}
K.~J. Liang, J.~B. Sigman, G.~P. Spell, D.~Strellis, W.~Chang, F.~Liu,
  T.~Mehta, and L.~Carin, ``{Toward Automatic Threat Recognition for Airport
  X-ray Baggage Screening with Deep Convolutional Object Detection},'' {\em
  arXiv preprint arXiv:1912.06329} , 2019.

\bibitem{Simonyan2015}
K.~Simonyan and A.~Zisserman, ``{Very Deep Convolutional Networks for
  Large-Scale Image Recognition},'' in {\em In Proceedings of the International
  Conference on Learning Representations (ICLR)},  May 2015.

\bibitem{inception_v2}
S.~Ioffe and C.~Szegedy, ``{Batch Normalization: Accelerating Deep Network
  Training by Reducing Internal Covariate Shift},'' in {\em {In Proccedings of
  the International Conference on Machine Learning (ICML)}},  2015.

\bibitem{he2016deep}
K.~He, X.~Zhang, S.~Ren, and J.~Sun, ``{Deep Residual Learning for Image
  Recognition},'' in {\em In Proceedings of the IEEE Conference on Computer
  Vision and Pattern Recognition (CVPR)},  June 2016.

\bibitem{Dai2016}
J.~Dai, Y.~Li, K.~He, and J.~Sun, ``{R-FCN: Object Detection via Region-based
  Fully Convolutional Networks},'' {\em Advances In Neural Information
  Processing Systems} , 2016.

\bibitem{Redmon2017}
J.~Redmon and A.~Farhadi, ``{YOLO9000: Better, Faster, Stronger},'' {\em
  Conference on Computer Vision and Pattern Recognition} , 2017.

\bibitem{Akcay2020}
S.~Akcay and T.~Breckon, ``{Towards Automatic Threat Detection: A Survey of
  Advances of Deep Learning within X-ray Security Imaging},'' {\em arXiv
  preprint arXiv:2001.01293} , 2020.

\bibitem{Bhowmik2019}
N.~Bhowmik, Q.~Wang, Y.~F.~A. Gaus, M.~Szarek, and T.~P. Breckon, ``{The Good,
  the Bad and the Ugly: Evaluating Convolutional Neural Networks for Prohibited
  Item Detection Using Real and Synthetically Composited X-ray Imagery},'' {\em
  arXiv preprint arXiv:1909.11508} , 2019.

\bibitem{yang2020object}
Y.~Yang, K.~J. Liang, and L.~Carin, ``Object detection as a positive-unlabeled
  problem,'' {\em arXiv preprint arXiv:2002.04672} , 2020.

\bibitem{Lin2017}
T.-Y. Lin, P.~Goyal, R.~Girshick, K.~He, and P.~Doll{\'{a}}r, ``{Focal Loss for
  Dense Object Detection},'' {\em International Conference on Computer Vision}
  , 2017.

\bibitem{Huang2017}
J.~Huang, V.~Rathod, C.~Sun, M.~Zhu, A.~Korattikara, A.~Fathi, I.~Fischer,
  Z.~Wojna, Y.~Song, S.~Guadarrama, and K.~Murphy, ``{Speed/accuracy trade-offs
  for modern convolutional object detectors},'' {\em Conference on Computer
  Vision and Pattern Recognition} , 2017.

\end{thebibliography}
\end{document}